\documentclass[letterpaper, 10 pt, conference]{ieeeconf}  

\IEEEoverridecommandlockouts                              

\usepackage{graphics} 
\usepackage{epsfig} 
\usepackage{mathptmx} 
\usepackage{times} 
\usepackage{amsmath} 
\usepackage{amssymb}  
\usepackage{subfigure}
\usepackage{multirow}
\usepackage{subcaption}
\usepackage{float}
\usepackage{ctable}
\usepackage{cuted}
\usepackage{colortbl}
\usepackage{algorithm}
\usepackage{algorithmic}
\usepackage{titlesec}
\usepackage{soul}
\usepackage{booktabs}

\usepackage[numbers,sort&compress]{natbib}
\usepackage{eucal} 
\usepackage[colorlinks,linkcolor=blue]{hyperref}
\title{\Large \bf
RESample: A \underline{R}obust Data Augmentation Framework via \underline{E}xploratory \underline{Sampling} for Robotic Manipulation
}

\author{\begin{tabular}{c}
    Yuquan Xue\textsuperscript{1}, Guanxing Lu\textsuperscript{2}, Zhenyu Wu\textsuperscript{3},  Chuanrui Zhang\textsuperscript{1}, Zhengyi Gu\textsuperscript{1}, Bofang Jia\textsuperscript{1}, Ziwei Wang\textsuperscript{1*}\\
  \end{tabular}

\thanks{$^{1}$Nanyang Technological University
$^{2}$Tsinghua University
$^{3}$Beijing University of Posts and Telecommunications
}
}

\begin{document}

\maketitle
\thispagestyle{empty}
\pagestyle{empty}







\begin{abstract}
Vision-Language-Action (VLA) models have shown strong manipulation capability when trained with large-scale imitation learning datasets.
However, these datasets that predominantly consist of successful trajectories rarely provide the corrective supervision required when execution deviates from standard demonstrations.
During deployment, these physical deviations lead to the distributional shift that drives policy to failure scenarios, yet missing failure recovery data prevents policy from correcting these execution deviations.
To address this failure recovery problem, we propose a coverage-guided data augmentation framework \textit{RESample} to actively supplement demonstration datasets for failure recovery. 
Specifically, to guide the augmentation and generate failure modes that possibly appear in the real world, RESample trains a conservative coverage function to identify failure cases that reside within the actual data distribution but are missing in the standard successful demonstrations.
Guided by the evaluated coverage discrepancy, we perform exploratory sampling to actively sample exploration behaviors followed by recovery actions, extending the coverage of training data with failure recovery trajectories.
With the augmented trajectory, the refined policy, which deviated in real settings, can recover from failure.
Experiments on the LIBERO benchmark and real-world manipulation tasks show that RESample consistently improves policy success rates, achieving up to 12\% absolute gain with no more than 20\% additional samples.

\end{abstract}

\section{INTRODUCTION}
\label{sec:intro}
\begin{figure}[t]
  \centering
  \includegraphics[width=1.0\linewidth]{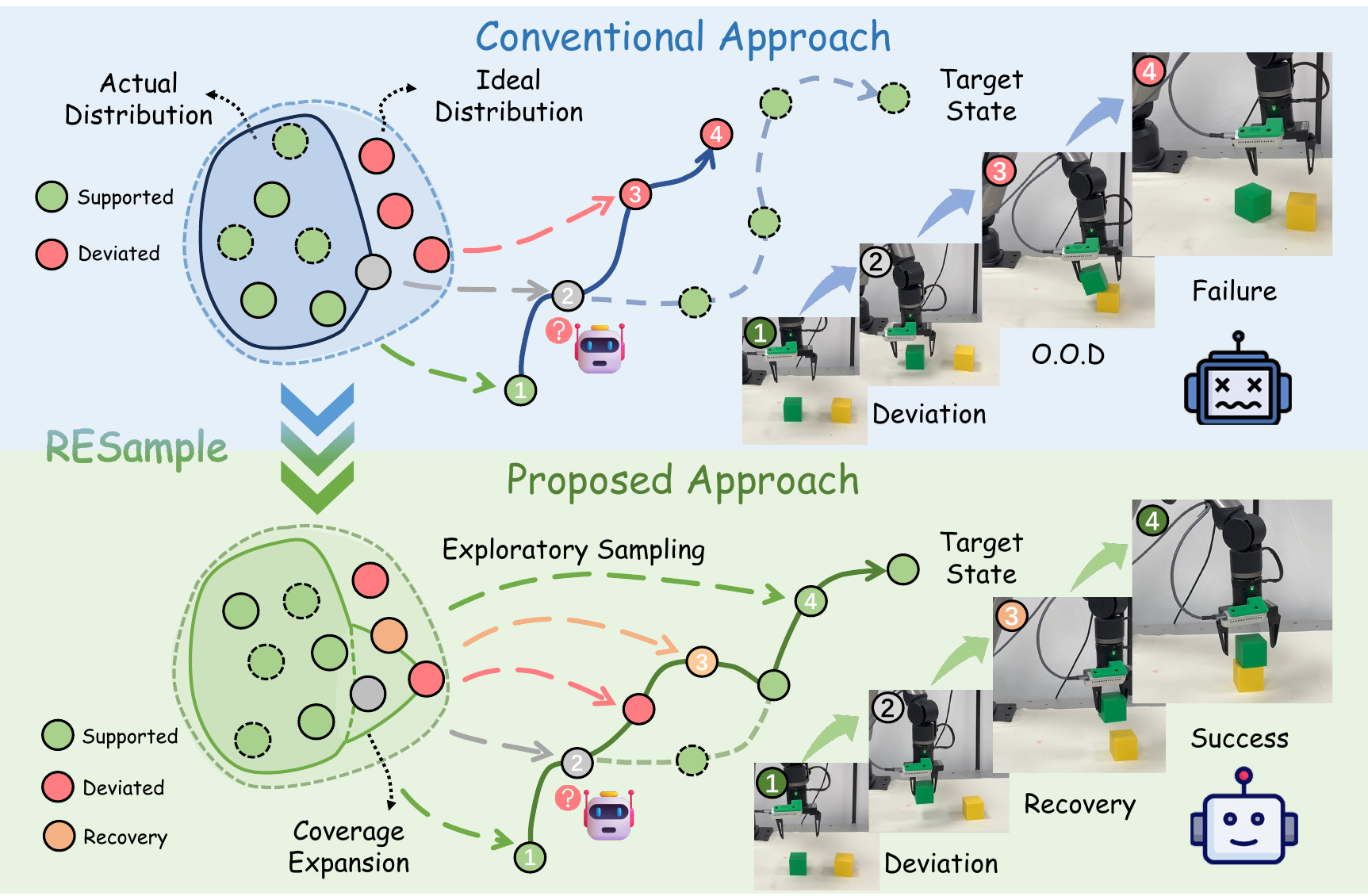}
  \caption{Illustration of the recovery-data gap in imitation learning.
  Top: Conventional policies trained on successful demonstrations may fail once small execution errors move the robot away from the demonstrated path.
  Bottom: RESample augments the dataset with targeted deviation-and-recovery trajectories, enabling the policy to correct common execution errors and complete the task.}
  \label{fig:teaser}
  \vspace{-0.7cm}
\end{figure}

Recent advances in Vision-Language-Action~(VLA) models~\cite{zitkovich2023rt,team2024octo,o2024open,kim2024openvla,black2024pi_0, kim2025finetuning} have enabled robots to learn complex manipulation skills from large-scale demonstrations.
Most of these models are trained through imitation learning, where the policy learns to reproduce continuous actions from expert demonstrations.
However, successful demonstrations alone do not provide supervision for failure recovery after execution deviations.
During deployment, physical perturbations such as gripper misalignment, unstable contact, object shifting, and slipping frequently drive the policy away from the demonstrated execution, leading to failure scenarios that require corrective actions.
Unfortunately, current large-scale imitation learning datasets~\cite{o2024open, walke2024bridge} predominantly contain successful trajectories and rarely include such recovery behaviors.
As a result, once the policy deviates from the demonstrated path, it hardly receives supervision on how to return to successful execution, causing errors to accumulate into task failure.
As illustrated in Fig.~\ref{fig:teaser}, the fundamental limitation is therefore the absence of failure recovery data for common execution deviations.

Existing approaches only partially address this problem.
From the algorithmic perspective, reinforcement learning-based methods can improve policy robustness by introducing online interaction with the environment~\cite{lu2025vla, zhang2025reinbot}.
However, such interaction is often sample-inefficient, unstable, and expensive for robotic manipulation, especially when large VLA policies are involved~\cite{chen2025conrft, lei2025rl}.
From the data perspective, methods including model-based data generation~\cite{xue2025demogen,mandlekar2023mimicgen} and data augmentation techniques~\cite{kostrikov2020image, peng2018sim,tobin2017domain} have been proposed to expand training datasets in scale and diversity utilizing agmentation methods including visual augmentations, target object texture changes and environment background changes.
While these methods improve the generalization against appearance variation or expand training data, they do not explicitly supplement the failure recovery behaviors that are weakly represented in successful demonstrations.
Furthermore, simply expanding the demonstration data scale without proper guidance and selection often leads to higher costs and inefficient usage of data, and random perturbations may even introduce uninformative or misleading supervision.

To address this failure recovery problem, we propose \textbf{RESample}, a guided data augmentation framework that actively supplements demonstration datasets with policy-induced failure recovery trajectories.
The key idea is to leverage the learned policy itself to expose execution deviations that are likely to occur in deployment, and then augment the dataset with corresponding recovery behaviors that return these deviations toward demonstrated successful execution.
Unlike conventional imitation learning, which only supervises nominal task execution, RESample provides explicit supervision on how to recover from realistic deployment failures.
failure cases that reside within the actual data distribution but are missing in the standard successful demonstrations.
To guide the data augmentation process, RESample trains a coverage function to identify the potential failure cases out of successful demonstrations, thereby evaluating the discrepancy between the demonstrated data distribution and the behaviors encountered during policy rollout.
Guided by the estimated coverage discrepancy, RESample performs exploratory sampling to collect policy-induced trajectories that actively sample execution errors and subsequently correct toward demonstrated successful behaviors.
These failure recovery trajectories are then incorporated into the original demonstrations to expand the coverage of the training data with targeted failure recovery supervision.
To validate the effectiveness of our method, we conduct extensive experiments on the LIBERO benchmark~\cite{liu2023libero} and a series of real-world manipulation tasks.
RESample improves success rates across multiple policy backbones, achieving up to \textbf{12\%} absolute improvement with no more than 20\% additional samples.

\section{RELATED WORKS}
\label{sec:relatedwork}
\subsection{Large Vision-Language-Action Models}

Recent progress in Vision-Language-Action~(VLA) models has established a powerful paradigm for learning generalist robotic manipulation policies.
These models combine visual observations, proprioceptive states, and language instructions to predict low-level action sequences~\cite{shao2025large}.
OpenVLA~\cite{kim2024openvla}, OpenVLA-OFT~\cite{kim2025finetuning}, $\pi_0$-FAST~\cite{pertsch2025fast} and VLA-Adapter~\cite{wang2025vlaadapter} introduced action tokenization for autoregressive policy learning, while Octo~\cite{team2024octo} and RDT-1B~\cite{liu2024rdt} explored diffusion-based~\cite{chi2023diffusion} continuous action denoising.
Recent $\pi_0$ and $\pi_{0.5}$ models~\cite{black2024pi_0,intelligence2025pi05} further improve action generation with flow-matching architectures and large vision-language backbones.

Despite these advances, most VLA models still rely heavily on behavior cloning from static successful demonstrations.
Such demonstrations provide dense supervision for nominal task execution, but they contain limited corrective behavior after the robot deviates from the demonstrated path~\cite{levine2020offline}.
During deployment, small errors in grasping, contact, alignment, or placement can therefore move the policy into states where the training data provides little guidance.
This limitation is especially important for long-horizon and contact-rich manipulation, where early execution errors can accumulate into task failure~\cite{lu2023imitation,de2019causal}.
Our work addresses this limitation from the data side by augmenting successful demonstrations with targeted deviation-and-recovery trajectories.

\subsection{Data Augmentation for Imitation Learning}

Data augmentation has been widely studied as a way to reduce the data dependency of imitation learning.
Existing approaches can be broadly grouped into heuristic-based augmentation and learning-based data generation.
Heuristic-based methods apply predefined transformations, including visual perturbations commonly used in self-supervised learning~\cite{laskin2020reinforcement,kostrikov2020image} and domain randomization in simulation~\cite{tobin2017domain,peng2018sim,akkaya2019solving}.
These methods are computationally efficient and can improve robustness to appearance changes, but they mainly alter observations or the environment's appearance.
Consequently, such augmentations do not directly provide corrective actions for physical execution deviations, which are central to recovery in manipulation tasks.

Learning-based methods synthesize additional robot data through more expressive mechanisms.
Reinforcement learning can explore new trajectories through online interaction~\cite{lu2025vla, zhang2024grape, zhang2025reinbot}, while model-based and generative approaches can create synthetic demonstrations~\cite{mandlekar2023mimicgen, xue2025demogen}.
Although these methods can expand the range of training experiences, online interaction may be sample-inefficient and costly for robotic systems~\cite{dulac2019challenges,amodei2016concrete}, while generated trajectories may suffer from sim-to-real gaps, physical implausibility, or action-observation misalignment~\cite{mandlekar2023mimicgen}.
More importantly, increasing data volume alone does not guarantee that the added samples contain the missing recovery behaviors needed by the deployed policy.

RESample differs from these prior augmentation strategies by focusing on policy-induced recovery data.
Rather than applying generic visual transformations, collecting arbitrary additional rollouts, or relying on synthetic demonstrations, RESample uses the current policy to expose common execution deviations and a Coverage Function to select weakly represented but recoverable trajectories.
The resulting data specifically supplements successful demonstrations with corrective behaviors, which directly match the recovery-data gap identified in imitation learning.

\section{PROPOSED METHOD}
\label{sec:method}
\begin{figure*}[!t]
    \centering
    \vspace{-0.5cm}
    \includegraphics[width=1.0\textwidth]{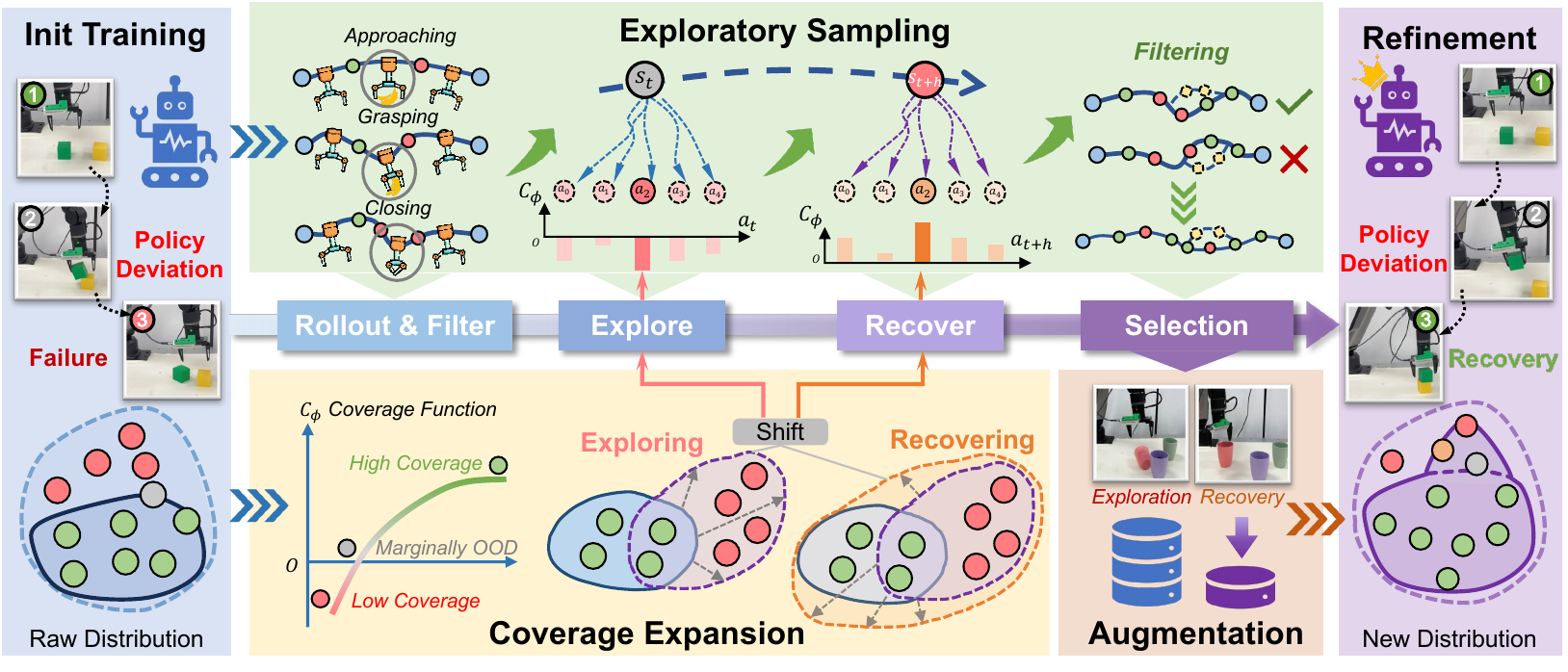}
    \caption{\textbf{An Overview of RESample Framework.}
    RESample first trains an initial policy and a coverage function on successful demonstrations.
    During policy rollout, interaction-critical states are detected, where candidate action chunks are sampled from the current policy.
    The coverage function then selects weakly covered but recoverable deviation trajectories, which are added to the original demonstrations to refine the policy with targeted recovery data.}
    \label{fig:pipeline}
    \vspace{-0.5cm}
\end{figure*}

\subsection{Problem Statement}
\label{sec:problem}


As discussed in Sec.~\ref{sec:intro}, the fundamental limitation of standard imitation learning is not only the finite size of demonstration datasets, but also the lack of recovery supervision after execution deviations.
We formulate this limitation as a data sampling problem.

Let $x=(s, a)$ denote a state-action pair.
The successful demonstration dataset $D_{\mathrm{exp}}$ is regarded as a finite sample drawn from the underlying successful manipulation distribution $p_{\star}(x)$, while the current policy induces its own deployment distribution $p_{\theta}(x)$. 
The objective of RESample is therefore to reduce the discrepancy between the empirical demonstrations and the policy rollout distribution by actively supplementing missing recovery behaviors.
Since only sampled demonstrations are available, we formulate this discrepancy as a two-sample problem and instantiate it using Maximum Mean Discrepancy (MMD):
\begin{equation}
\pi_\theta^\star
=
\arg\min_{\pi_\theta}
\operatorname{MMD}
\!\left(
p_{\star},\,
p_\theta
\right),
\label{eq:objective}
\end{equation}
where MMD measures the discrepancy between the empirical demonstration distribution and the policy rollout distribution through pairwise sample similarity without requiring explicit density estimation.



RESample addresses this discrepancy by augmenting the original demonstrations $D_{\mathrm{exp}}$ with policy-induced recovery trajectories $D_{\mathrm{res}}$.
Since these additional trajectories are sampled from policy rollout while complementing underrepresented recovery behaviors, the resulting augmented dataset provides a closer empirical approximation to the policy rollout distribution than the original demonstrations alone.
Consequently, RESample reduces the distribution discrepancy through guided data augmentation, enabling the augmented demonstrations to better match the behaviors encountered during deployment.



\subsection{RESample Framework Overview}
\label{sec:overview}


RESample is a coverage-guided data augmentation framework that actively supplements successful demonstrations with failure recovery trajectories.
The key idea is to leverage policy rollout to identify realistic execution deviations and augment the training data with failure recovery trajectories.
Fig.~\ref{fig:pipeline} illustrates this process.

Starting from the policy and coverage function trained on demonstration $D_{\mathrm{exp}}$, RESample repeatedly samples multiple candidate trajectories under the same state at interaction-critical stages of manipulation.
The coverage function is then used to guide filtering out the satisfying failure recovery samples.
The selected trajectories are collected into a recovery dataset $D_{\mathrm{res}}$, which is merged with the original demonstrations to form the augmented dataset $D_{\mathrm{aug}}$.
Consequently, the augmented data captures realistic deployment-time execution deviations while explicitly providing the missing recovery absent from successful demonstrations.

\subsection{Coverage Function Training}
\label{sec:coverage}

To guide data augmentation, RESample first estimates how well a rollout behavior is represented by the successful demonstrations.
For a state-action pair $x=(s,a)$, its demonstration support is defined as
\begin{equation}
S(x)
=
\mathbb{E}_{x'\sim D_{\mathrm{exp}}}
k(x,x'),
\label{eq:kernel_support}
\end{equation}
where $k(\cdot,\cdot)$ denotes the kernel similarity used in the MMD formulation.
Larger support indicates that the behavior is well represented by the demonstrations, whereas smaller support corresponds to underrepresented regions of the empirical dataset.

Directly evaluating Eq.~\eqref{eq:kernel_support} for every rollout sample is intractable in the high-dimensional visual state-action space.
Following the common practice of parameterizing kernel-based distribution representations with neural functions~\cite{Muandet_2017,kim2018imitation}, we parameterize the demonstration support with a neural Coverage Function $C_\phi(x)$, which serves as a practical estimator of demonstration support and guides the exploratory sampling process.


The coverage function is implemented as a conservative support value function that estimates whether a state-action pair is supported by the demonstrations and can continue toward demonstrated success.
It is trained using an in-distribution temporal objective together with a conservative regularization term:
\begin{equation}
\begin{aligned}
    \mathcal{L}_{\mathrm{C}}(\phi)
    =\;&
    \mathbb{E}
    \left[
    \left(
    C_\phi(s,a)-y(s')
    \right)^2
    \right]
    \\
    &+
    \alpha\,
    \mathbb{E}
    \left[
    \mathbb{E}_{\tilde a\sim\pi_\theta}
    C_\phi(s,\tilde a)
    -
    \mathbb{E}_{a\sim D_{\mathrm{exp}}}
    C_\phi(s,a)
    \right]
\end{aligned}
\label{eq:coverage_training}
\end{equation}
where
$y(s')=1+\gamma\mathbb{E}_{a'\sim\pi_\theta}C_{\bar{\phi}}(s',a')$,
$C_{\bar{\phi}}$ is a target network, and $\alpha$ controls the strength of conservative regularization.
The training objective consists of two complementary components.
The first term propagates support along successful demonstrations, encouraging demonstrated behaviors and their successful continuations to receive consistently high coverage values.
The second term suppresses policy-sampled actions that are weakly supported by the demonstrations, preventing unsupported rollout behaviors from receiving artificially high coverage values.
Together, these two objectives learn a conservative estimator of demonstration support, assigning high coverage to demonstrated behaviors while suppressing unsupported policy behaviors.
The resulting coverage function provides the guidance signal for the exploratory sampling procedure described in Sec.~\ref{sec:mechanism}.



\subsection{RESample Algorithm}
\label{sec:sampling}

Given the learned Coverage Function, RESample constructs the recovery dataset $D_{\mathrm{res}}$ through an exploratory sampling process, as summarized in Alg.~\ref{alg:resample}.
During rollout, the method first detects interaction-critical stages of manipulation, repeatedly samples candidate trajectories from the current policy, and selects an informative recovery trajectory according to the sampling mechanism described in Sec.~\ref{sec:mechanism}.
The selected trajectories are collected into $D_{\mathrm{res}}$ and used together with the original demonstrations to refine the policy.


\begin{algorithm}[t]
\caption{RESample: Exploratory Sampling}
\label{alg:resample}
\small
\begin{algorithmic}[1]
\REQUIRE Dataset $D_{\mathrm{exp}}$, policy $\pi_\theta$, Coverage Function $C_\phi$
\STATE $D_{\mathrm{res}}\leftarrow\emptyset$
\FOR{each state $s_t$}
    \STATE $\eta_t\leftarrow f_{\mathrm{int}}(s_t,\Delta x_t,g_t)$
    \IF{$\eta_t>\rho_{\mathrm{int}}$}
        \STATE $\{\tau^{(k)}\}_{k=1}^{K}\sim\pi_\theta(\cdot|s_t)$
        \STATE $\tau^\ast\leftarrow\textsc{ExploratorySampling}
        (\{\tau^{(k)}\}_{k=1}^{K},C_\phi)$
        \IF{$\mathcal{V}(\tau^\ast)=1$}
            \STATE $D_{\mathrm{res}}
            \leftarrow
            D_{\mathrm{res}}\cup\{\tau^\ast\}$
        \ENDIF
    \ENDIF
\ENDFOR
\STATE Refine $\pi_\theta$ on $D_{\mathrm{exp}}\cup D_{\mathrm{res}}$
\RETURN $\pi_\theta$
\end{algorithmic}
\end{algorithm}

\textbf{RESample Triggering.}
Rather than sampling uniformly throughout the rollout, RESample activates exploratory sampling only during interaction-critical stages of manipulation, including grasping, contact, alignment, and placement, where small execution deviations are most likely to require corrective actions~\cite{zhao2023act,kanehira2025rl}. 
In practice, these stages are detected using simple robot motion and gripper cues: slow EEF motion together with active gripper engagement reliably indicates physical interaction with the manipulated object.



\textbf{Candidate Generation.}
Once an interaction-critical state is detected, RESample samples $K$ candidate trajectories from the same initial state.
Since all candidates are generated by the deployed policy, they naturally capture realistic execution deviations encountered during deployment.
The coverage function then evaluates these candidates and selects the most informative recovery trajectory according to the exploratory sampling mechanism.

\textbf{Deviation Discovery.}
To discover failure recovery cases, the first part of a candidate trajectory should leave well-covered demonstrated behaviors and enter regions receiving weak support from the demonstrations.
Since the coverage function estimates demonstration support, lower coverage naturally indicates behaviors that are rarely or never observed in the successful dataset.
Therefore, we measure the deviation score using the average coverage over the exploratory segment,
\begin{equation}
\mathcal{J}_{\mathrm{dev}}(\tau)
=
-
\frac{1}{H_e}
\sum_{h=0}^{H_e-1}
C_\phi(s_{t+h},a_{t+h}),
\label{eq:deviation_score}
\end{equation}
where $H_e$ denotes the length of the exploratory segment.
Using the average coverage over the exploratory segment measures the overall extent of the deviation rather than isolated low-coverage states.
Consequently, maximizing $\mathcal{J}_{\mathrm{dev}}$ encourages trajectories that consistently explore underrepresented behaviors while avoiding candidates whose deviations occur only momentarily.

\textbf{Recovery Evaluation.}
However, exploration alone is insufficient for failure recovery learning.
Aggressive exploration may easily produce irreversible failures that cannot provide useful supervision for policy improvement.
Instead, a desirable recovery trajectory should gradually return toward behaviors well supported by the demonstrations, indicating that the deviation remains recoverable.
Therefore, recoverability is evaluated on the remaining horizon,
\begin{equation}
\mathcal{J}_{\mathrm{rec}}(\tau)
=
\frac{1}{H-H_e}
\sum_{h=H_e}^{H-1}
C_\phi(s_{t+h},a_{t+h}),
\label{eq:recovery_score}
\end{equation}
where $H$ denotes the whole trajectory horizon.
A larger recovery score indicates that the trajectory successfully reconnects to the demonstrated behavior distribution after exploration.
Using the average coverage over the recovery segment measures the quality of the entire recovery process rather than the final state alone, encouraging smooth corrective behaviors instead of accidental recovery near the trajectory end.

\textbf{Candidate Selection.}
The final trajectory is selected by jointly maximizing the two complementary objectives:
\begin{equation}
\tau^\ast
=
\arg\max_{\tau}
\min
\left(
\mathcal{J}_{\mathrm{dev}}(\tau),
\mathcal{J}_{\mathrm{rec}}(\tau)
\right).
\label{eq:trajectory_selection}
\end{equation}
The two objectives play complementary roles.
$\mathcal{J}_{\mathrm{dev}}$ encourages the trajectory to explore informative execution deviations that are uncovered in the demonstrations, while $\mathcal{J}_{\mathrm{rec}}$ ensures that these deviations remain recoverable by returning toward demonstrated behaviors.
Instead of favoring trajectories with a high score in only one objective, the proposed objective maximizes the lower of the two scores, requiring both sufficient exploration and successful recovery.
Consequently, trajectories with aggressive but unrecoverable failures or trajectories that remain too close to the demonstrations are both discarded.
The selected trajectories therefore simultaneously expand the coverage of the demonstration dataset and provide meaningful recovery supervision.

\textbf{Data Augmentation.}
Each selected trajectory is first filtered by a validity check $\mathcal{V}(\tau)$ to ensure that it remains executable and reaches a recoverable segment.
The filtered recovery dataset is then merged with the original demonstrations to refine the policy.

\begin{table*}[t]
    \small
    \centering
    \caption{Experimental Results on LIBERO.}
    \setlength{\tabcolsep}{19.5pt}
    \begin{tabular}{clccccc}
    \toprule
    \multicolumn{7}{c}{\textbf{LIBERO}} \\
    \midrule
    \textbf{Size} & \textbf{Method} & \textbf{Spatial} & \textbf{Object} & \textbf{Goal} & \textbf{Long} & \textbf{Average}\\
    \midrule
    \multirow{3}{*}{\textit{Tiny}}
    & VLA-OS~\cite{}   & 87.0  & 96.5 & 92.7  & 66.0  & 85.6    \\
    & Diffusion Policy~\cite{chi2023diffusion}    & 68.5   & 93.5   & 72.5  & 73.5   & 77.0      \\
    & \textbf{Diffusion Policy + Ours} & \textbf{89.0} & \textbf{98.5} & \textbf{76.0} & \textbf{92.5} & \textbf{89.0} \\
    \midrule
    \multirow{4}{*}{\textit{Small}}
    & Octo~\cite{team2024octo}  & 78.9	& 85.7	& 84.6	& 51.1	& 75.1 \\
    & SmolVLA~\cite{}  & 93.0 & 94.0 & 91.0 & 77.0 & 88.8 \\
    & VLA-Adapter~\cite{wang2025vlaadapter}  & 91.2	& 93.2	& 97.4	& 90.8	& 93.1 \\
    & \textbf{VLA-Adapter + Ours} & \textbf{93.2} & \textbf{94.4} & \textbf{97.4} & \textbf{94.0} & \textbf{94.8} \\
    \midrule
    \multirow{4}{*}{\textit{Large}}
    & OpenVLA~\cite{o2024open}    & 84.7   & 88.4   & 79.2   & 53.7   & 76.5    \\
    & WorldVLA~\cite{}   & 87.6   & 96.2   & 83.4  & 60.0  & 81.8    \\
    & $\pi_0$~\cite{black2024pi_0}    & 95.2   & \textbf{97.6}  & 91.2  & 79.8  & 90.9    \\
    & \textbf{$\pi_0$ + Ours}    & \textbf{96.4}   & 96.4   & \textbf{91.4}  & \textbf{87.4}   & \textbf{92.9}   \\
    \bottomrule
    \end{tabular}
    \label{tab:exp_results}
    \vspace{-0.2cm}
\end{table*}

\begin{table*}[t]
\centering
\small
\caption{Real-World Experiment Results.}
\setlength{\tabcolsep}{16pt}
\begin{tabular}{lccccc}
\toprule
\textbf{Method} & \textbf{Pick Block} & \textbf{Stack Cup} & \textbf{Arrange Cubes} & \textbf{Stack 2 Cups} & \textbf{Average}\\
\midrule
Diffusion Policy        & 65.0   & 60.0   & 15.0   & 20.0 &  40.0    \\
\textbf{Diffusion Policy + Ours} & \textbf{80.0} & \textbf{70.0} & \textbf{30.0} & \textbf{45.0} & \textbf{56.3}  \\
\bottomrule
\end{tabular}
\label{tab:real_world_results}
\vspace{-0.4cm}
\end{table*}

\section{EXPERIMENTS}
\label{sec:exp}

We evaluate RESample from three perspectives that directly follow the motivation.
First, we test whether targeted deviation-and-recovery data improves policy success rates.
Second, we examine whether the effect holds across different policy backbones and model scales.
Third, we analyze whether the Coverage Function guides sampling toward weakly represented but recoverable behaviors.

Our evaluation includes LIBERO simulation tasks~\cite{liu2023libero} and real-world manipulation on a Galaxea A1 robot arm with a parallel gripper.
Success rate is used as the primary metric.

\subsection{Experimental Setup}

\textbf{Baseline Policies:}
For LIBERO, we apply RESample to three representative backbones, including Diffusion Policy~\cite{chi2023diffusion}, VLA-Adapter~\cite{wang2025vlaadapter}, and $\pi_0$~\cite{black2024pi_0}.
These policies differ in model scale and action modeling, which allows us to test whether RESample acts as a general data augmentation framework rather than a backbone-specific technique.
All policies take side-view images, wrist-view images, robot proprioception, and language instructions as input and output 7-DoF action chunks.
For real-world experiments, we use Diffusion Policy as the base policy.

\textbf{Training Details:}
All policies are trained with behavior cloning.
For Diffusion Policy, the initial policy is trained on each LIBERO suite for 50 epochs with batch size 64, AdamW, and a cosine learning-rate schedule initialized at $1\times10^{-4}$.
For VLA-Adapter, the policy is trained for 30k steps with a learning rate $1\times10^{-5}$.
For $\pi_0$, the base model is finetuned on LIBERO for 30k steps with batch size 256 and learning rate $2\times10^{-5}$.
After RESample augmentation, Diffusion Policy is retrained for another 50 epochs, while VLA-Adapter and $\pi_0$ are refined for another 10k steps.
The Coverage Function is trained separately for each task with a batch size of 256 and a learning rate of $1\times10^{-4}$.
All experiments use a single-round augmentation protocol, where valid exploration-recovery trajectories are retained and mixed with the original demonstrations.

\subsection{Experimental Results}

\begin{figure*}[t]
    \centering
    \includegraphics[width=1.0\linewidth]{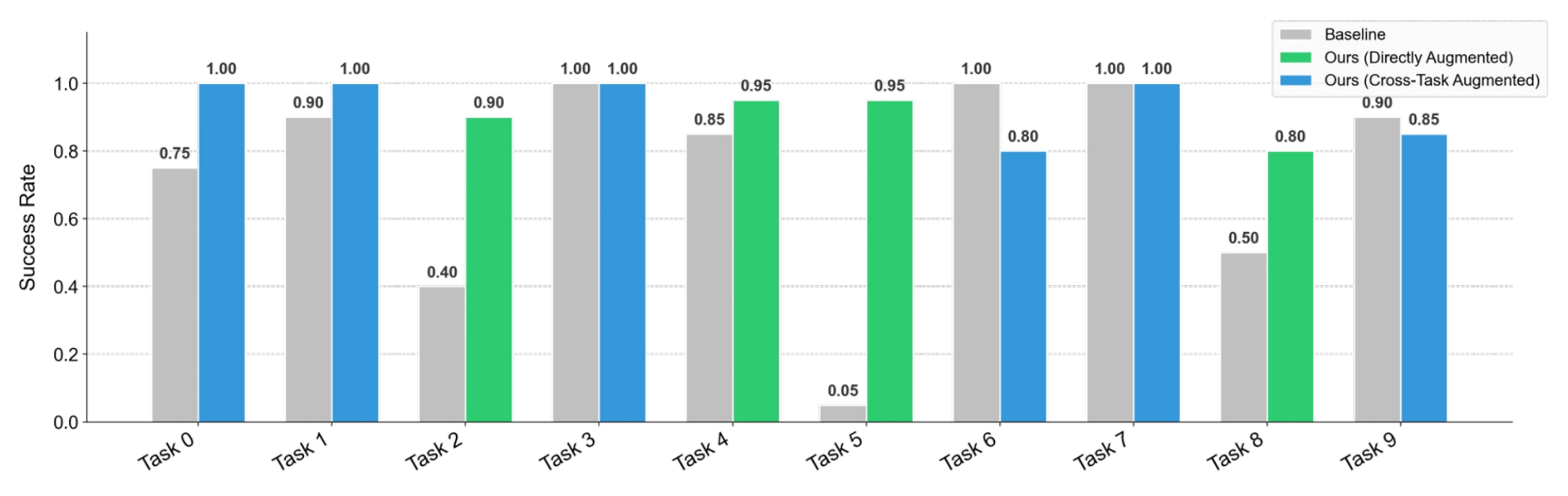}
    \caption{\textbf{Cross-Task Augmentation.} The augmented data generated from tasks {2, 4, 5, 8} can transfer to other tasks within the same category, leading to an additional performance boost of \textbf{5-10\%} on average.}
    \label{fig:cta}
    \vspace{-0.4cm}
\end{figure*}

\textbf{Simulation Results:}
Table~\ref{tab:exp_results} shows that RESample improves the corresponding base policies across model scales.
Diffusion Policy improves from 77.0\% to 89.0\%, VLA-Adapter improves from 93.1\% to 94.8\%, and $\pi_0$ improves from 90.9\% to 92.9\% on average.
These results support the role of RESample as a data augmentation procedure that can be applied to different policy architectures.

The most informative gains appear on the LIBERO-Long suite, where policies must pass through multiple interaction stages and early deviations can affect later execution.
Diffusion Policy improves from 73.5\% to 92.5\%, VLA-Adapter improves from 90.8\% to 94.0\%, and $\pi_0$ improves from 79.8\% to 87.4\%.
This trend matches our motivation because long-horizon tasks are more likely to require corrective behavior after small deviations.
RESample supplies such missing recovery data by sampling policy-induced deviations and retaining trajectories that return to well-supported behavior.

The gains are smaller on saturated suites where the base policy already performs strongly.
For example, VLA-Adapter remains at 97.4\% on Goal, while $\pi_0$ has limited room for improvement on Spatial and Object.
This behavior is expected because RESample is not designed as a generic accuracy booster.
The method mainly helps when the original demonstrations do not sufficiently cover the recovery behaviors needed by the deployed policy.

Fig.~\ref{fig:cta} further shows that RESample-generated data can transfer across tasks in the same category.
For example, recovery data generated from one spatial placement task can improve related spatial tasks with similar misalignment or occlusion patterns.
This suggests that the augmented trajectories capture reusable correction patterns rather than task-specific noise.

\begin{figure*}[t]
    \centering
    \includegraphics[width=1.0\linewidth]{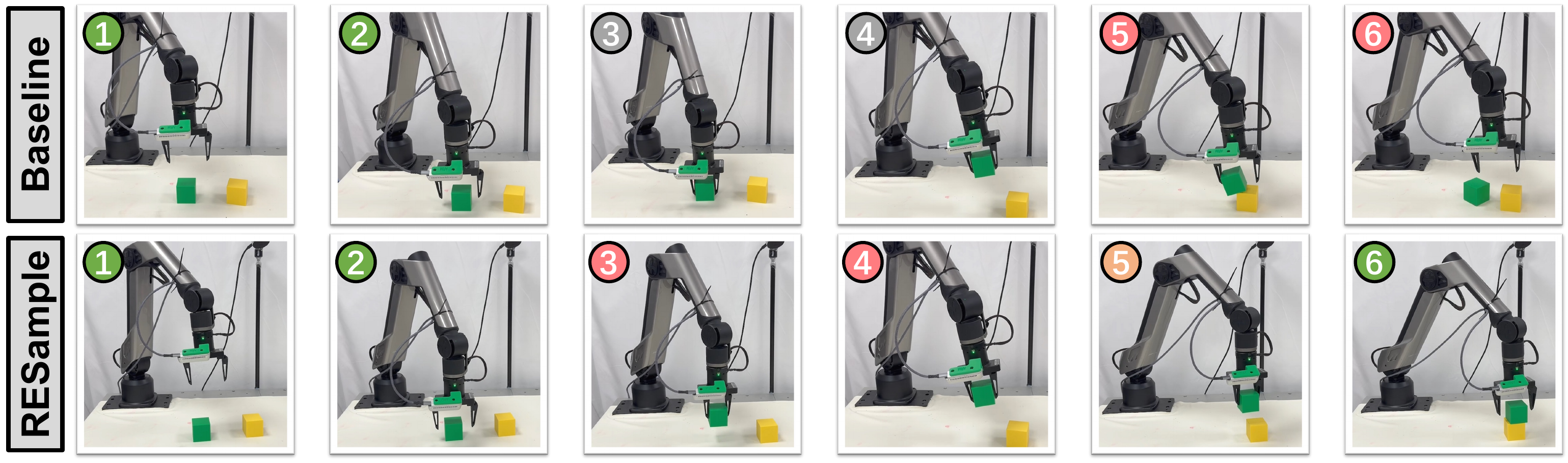}
    \caption{\textbf{Failure recovery comparison.}
    Compared with the baseline, which fails after a pose deviation, the refined policy recovers from the deviation and completes the manipulation task.
    }
    \label{fig:real_world}
    \vspace{-0.2cm}
\end{figure*}

\textbf{Real-World Results:}
Table~\ref{tab:real_world_results} reports results on \textit{Pick Block}, \textit{Stack Cup}, \textit{Arrange Cubes}, and \textit{Stack 2 Cups}.
RESample improves the average success rate from 40.0\% to 56.3\%.
The greatest improvements appear on tasks that require correction after object pose errors.
For \textit{Arrange Cubes}, the success rate improves from 15.0\% to 30.0\%, while \textit{Stack 2 Cups} improves from 20.0\% to 45.0\%.
These tasks expose the same missing-data issue described in the introduction, since clean demonstrations rarely cover the corrective motions needed after placement or stacking misalignment.

Fig.~\ref{fig:real_world} shows a representative recovery behavior in the \textit{Arrange Cubes} task.
The baseline policy fails when a small object misalignment disrupts the following placement stage.
After training via RESample, the policy observes similar deviation-and-recovery patterns during augmentation and learns to correct the cube pose before continuing the task.
The real-world results therefore confirm that the simulated gains come from improved recovery behavior rather than only benchmark-specific fitting.

\subsection{Ablation and Mechanism Analysis}

\begingroup
    \setlength{\tabcolsep}{10pt}
    \begin{table}[t]
    \centering
    \caption{Exploratory Sampling Ablation.}
    {\small
    \begin{tabular}{lc}
    \toprule
    \textbf{Ablation} & \textbf{LIBERO-Spatial}\\
    \midrule
    Base Policy        & 68.5  \\
    Naive Rollout Augmentation     & 74.8   \\
    Random Sampling Augmentation      & 73.0   \\
    \textbf{Ours}       & \textbf{76.5}   \\
    \bottomrule
    \end{tabular}
    }
    \label{tab:ablation_results}
    \vspace{-0.1cm}
    \end{table}
\endgroup
\begin{figure}[t]
    \centering
    \includegraphics[width=1.0\linewidth]{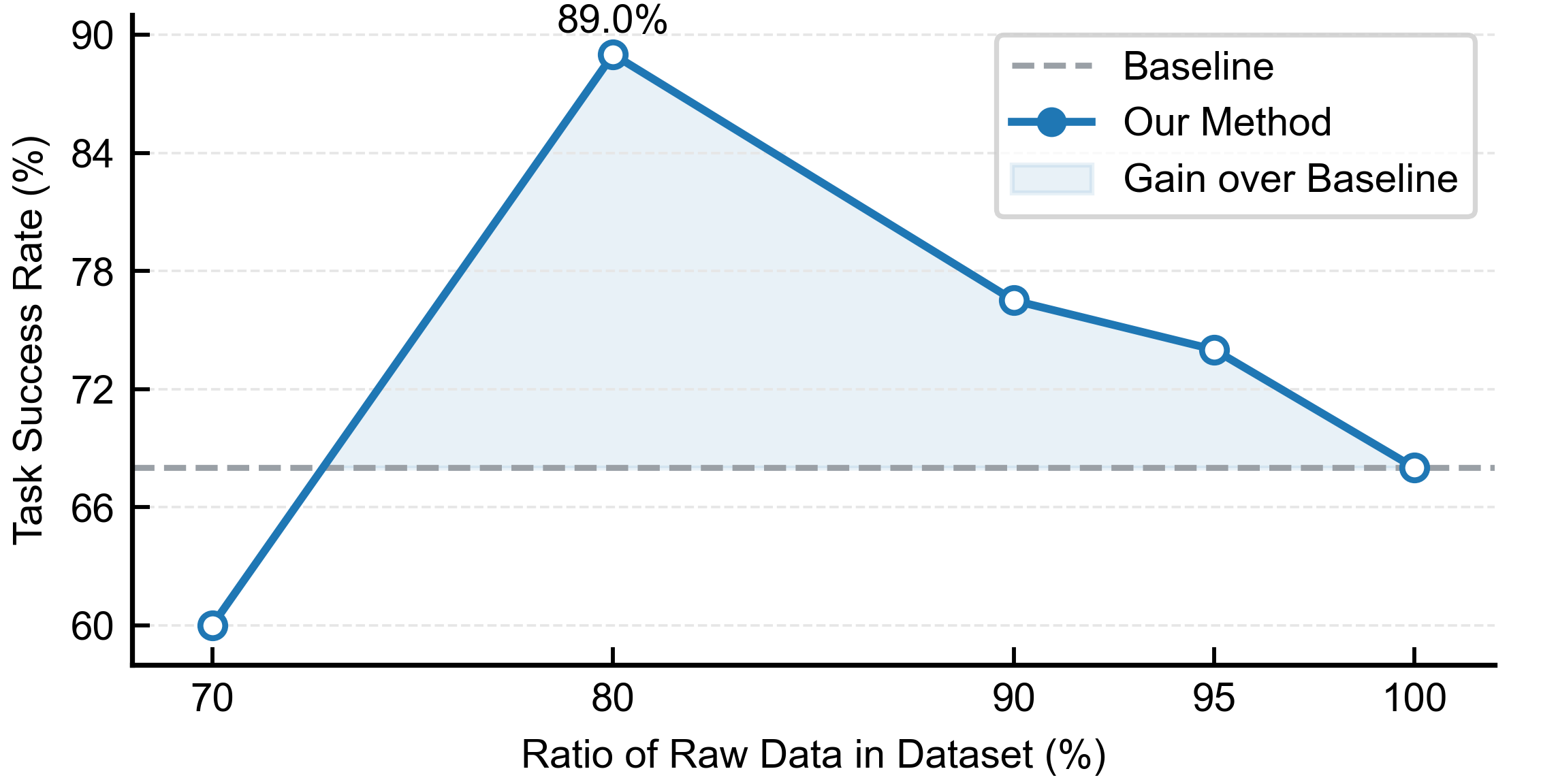}
    \caption{\textbf{Effect of Augmented Data Ratio.} RESample performs best in moderate mixing.}
    \label{fig:mix_ratio}
    \vspace{-0.6cm}
\end{figure}
\textbf{Ablation Study:}
Table~\ref{tab:ablation_results} isolates the contribution of coverage-guided exploratory sampling on LIBERO-Spatial with Diffusion Policy.
\textit{Base Policy} uses only original demonstrations.
\textit{Naive Rollout Augmentation} adds trajectories collected from initial-policy rollouts without targeted sampling.
\textit{Random Sampling Augmentation} injects random actions during rollout.
\textit{RESample} uses policy-induced sampling and Coverage Function guided trajectory selection.
All augmentation variants use the same additional data ratio.

Naive rollout augmentation improves over the base policy, which indicates that rollout data can add useful supervision.
Random sampling is less effective, suggesting that arbitrary perturbations are less aligned with the deployed policy's common deviations.
RESample achieves the best result because the selected samples both expose weakly represented behaviors and remain recoverable.
This comparison supports the central design choice that recovery augmentation should be targeted to policy-induced deviations rather than collected through untargeted data expansion.

Fig.~\ref{fig:mix_ratio} studies the ratio between raw demonstrations and RESample-generated data.
A moderate amount of augmented data gives the best performance, with the peak at 20\% augmented data in the final training set.
Too little augmented data provides limited recovery supervision, while too much augmented data may dilute the clean successful demonstrations.
RESample should therefore be viewed as a targeted supplement to successful demonstrations rather than a replacement for them.

\begin{figure}[t]
    \centering
    \includegraphics[width=1.0\linewidth]{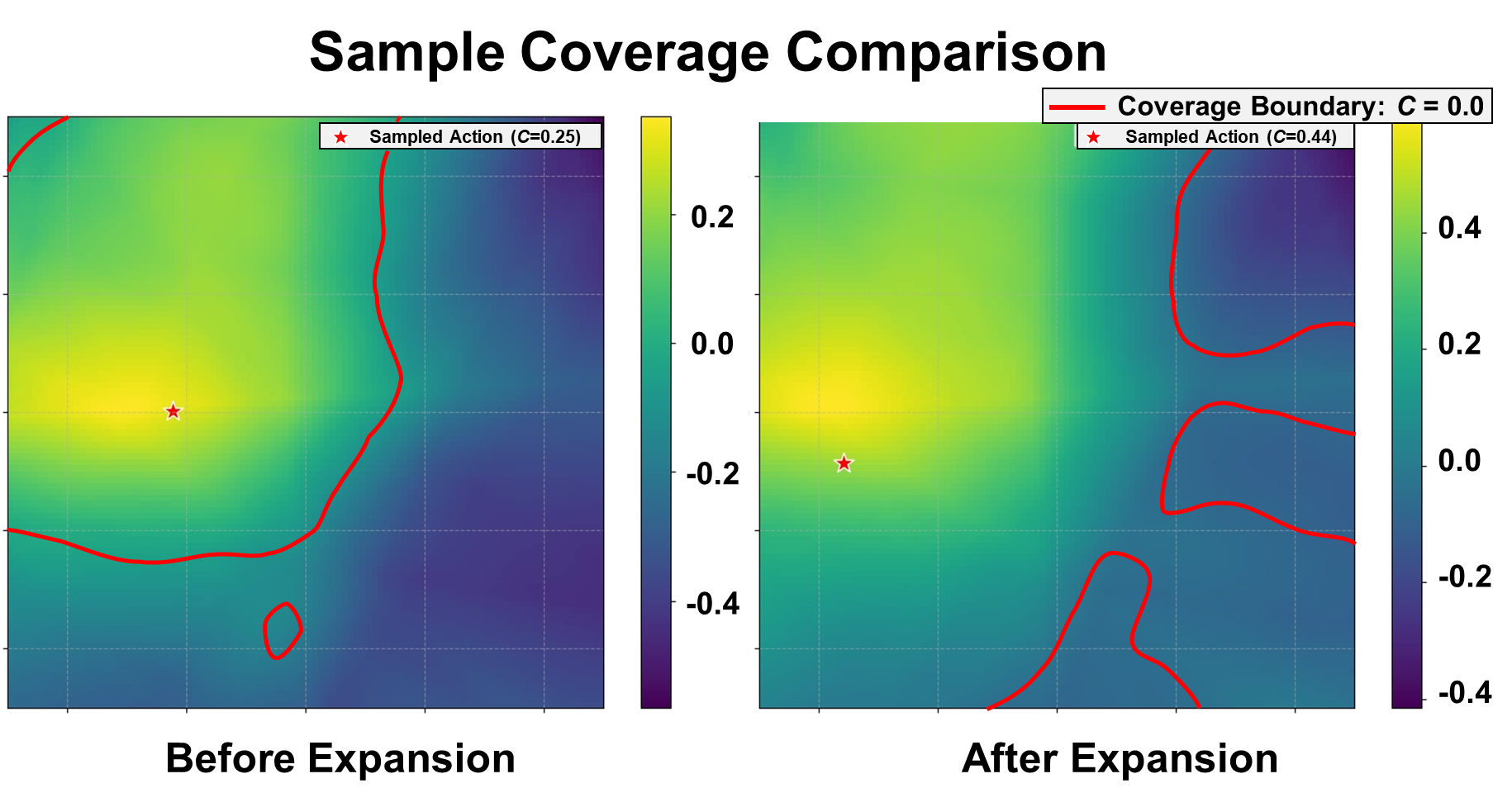}
    \caption{\textbf{Coverage Value Map.} Coverage value map before and after RESample augmentation. The augmented policy selects actions with higher coverage, while the updated dataset expands the supported region around interaction-critical states.}
    \label{fig:sample}
    \vspace{-0.6cm}
\end{figure}

\begin{figure}[t]
    \centering
    \includegraphics[width=1.0\linewidth]{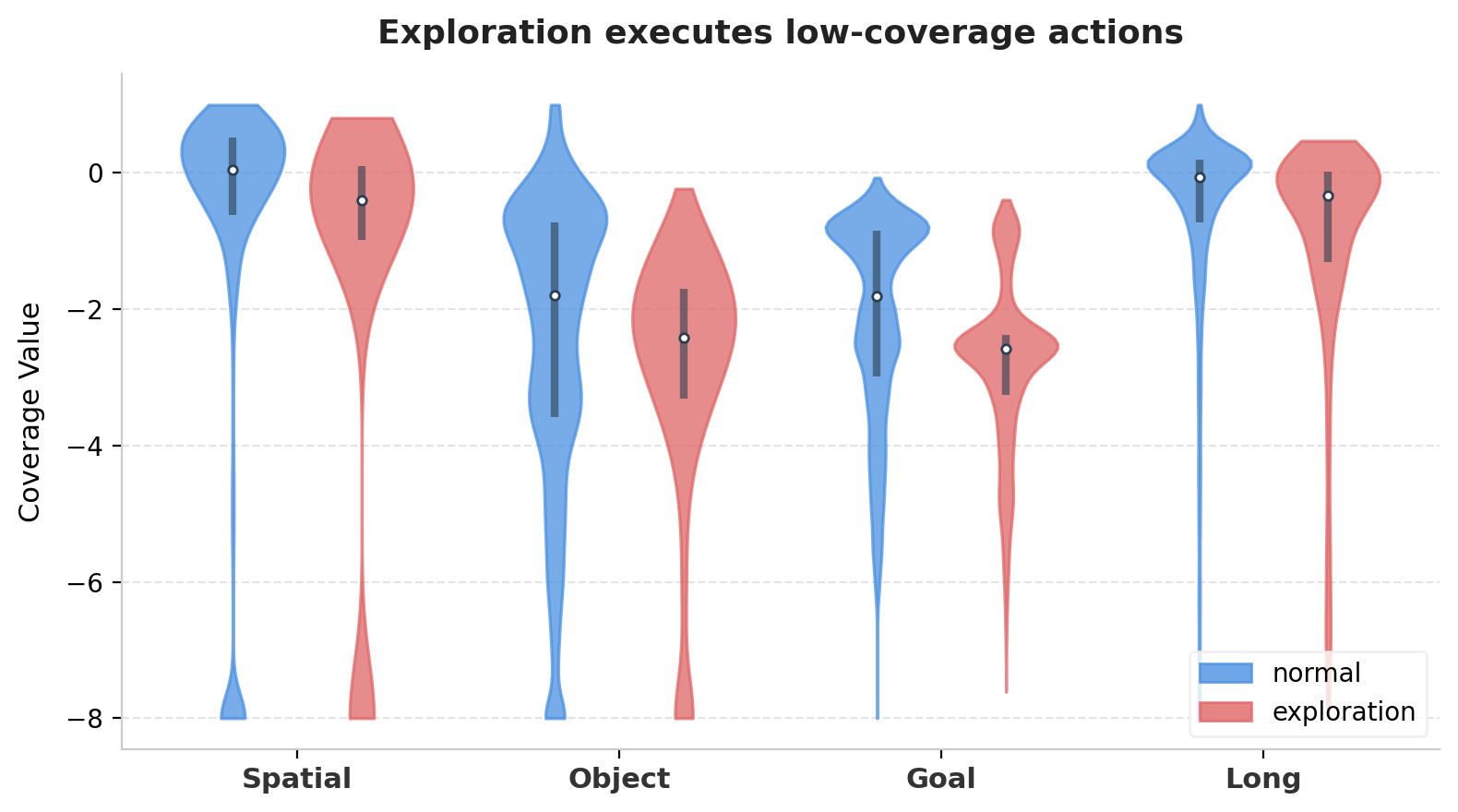}
    \caption{\textbf{Coverage Separation Comparison.}
    RESample-inserted actions receive lower coverage values, showing that RESample selects exploration actions.}
    \label{fig:qsep}
\end{figure}

\begin{figure}[t]
    \centering
    \includegraphics[width=1.0\linewidth]{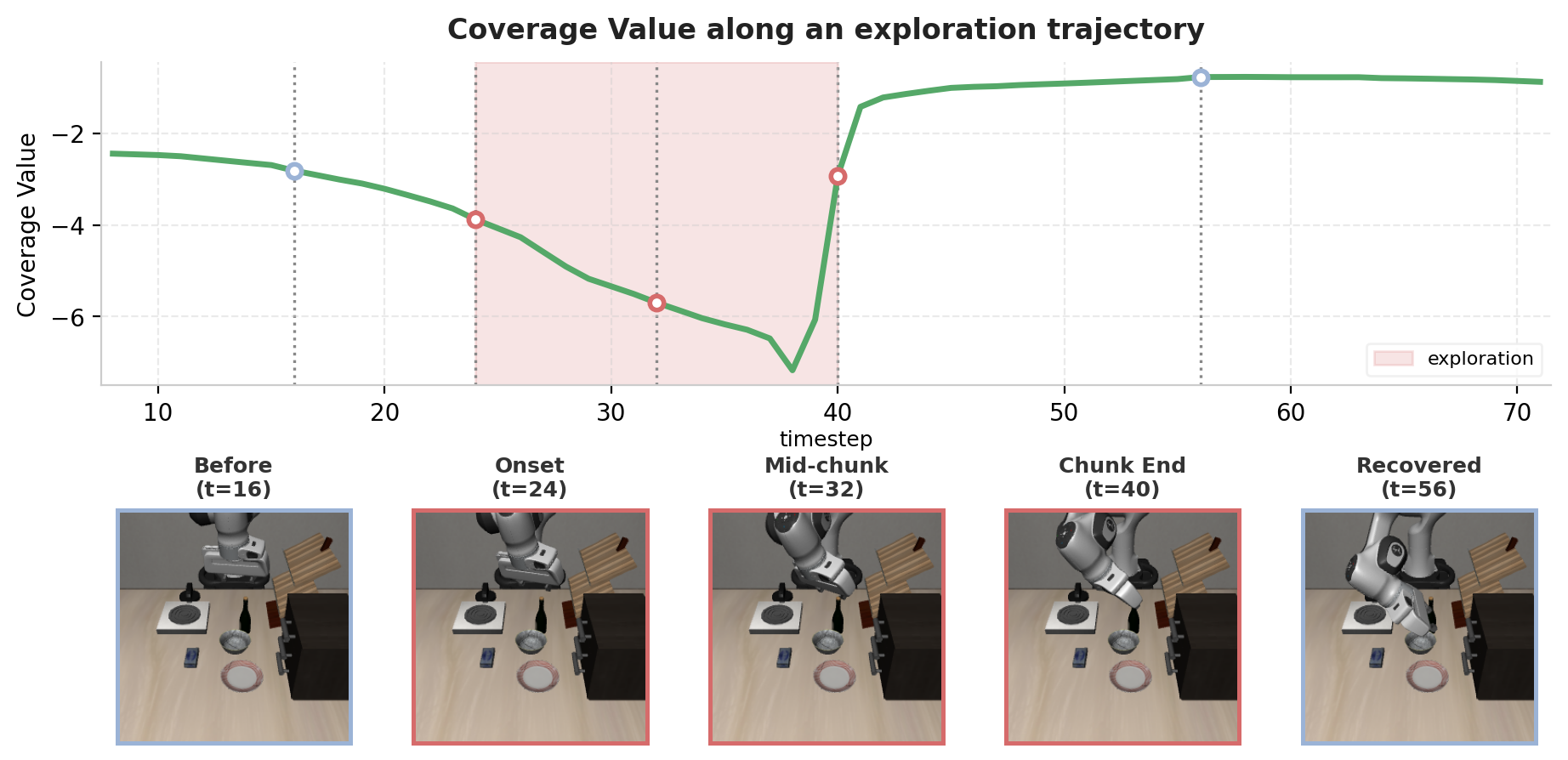}
    \caption{\textbf{Coverage Dynamics Along Timestep.}
    The coverage value decreases after exploration starts and then increases after the inserted chunk, showing a deviation-and-recovery pattern.}
    \label{fig:trajectory}
    \vspace{-0.6cm}
\end{figure}

\textbf{Mechanism Analysis:}
Fig.~\ref{fig:sample} visualizes the coverage value map before and after RESample augmentation.
After augmentation, the supported region expands around policy-relevant actions, and the retrained policy is encouraged to select actions with higher coverage near interaction-critical states.
This result shows that the augmented data changes the support of the training set rather than simply increasing its size.

Fig.~\ref{fig:qsep} compares Coverage Function values of normal policy actions and RESample-inserted actions.
Inserted actions receive lower coverage values, which confirms that RESample selects behaviors that are weakly represented by the original demonstrations.
Because these actions are sampled from the current policy rather than random noise, they correspond to plausible deployment deviations.

Fig.~\ref{fig:trajectory} further shows the temporal coverage pattern along a representative RESample trajectory.
Coverage drops during the inserted exploration chunk and rises after the recovery segment.
This pattern directly matches the intended deviation-and-recovery structure, where RESample first exposes a missing recovery case and then records how the policy returns toward well-supported behavior.

\section{CONCLUSION}
\label{sec:con}
This paper introduced RESample, a coverage-guided data augmentation framework for improving failure recovery in robotic manipulation. 
Instead of indiscriminately increasing demonstration data, RESample exposes base policy to interaction-critical deviations and a learned Coverage Function to select recoverable trajectories that are weakly represented in the original dataset. 
By augmenting demonstrations with these targeted exploration-recovery samples, RESample provides policies with missing corrective supervision and improves robustness across both LIBERO and real-world manipulation tasks. 
The foundamental limitation is that the method meets a performance bottleneck in saturated tasks.
Future work will explore online coverage updates and curriculum-based sampling to progressively expand recovery coverage in more diverse manipulation scenarios.

\begingroup
\renewcommand{\baselinestretch}{0.98} 
\small 
\setlength{\bibsep}{0.2pt}    
\bibliographystyle{IEEEtranN}
\bibliography{references}
\endgroup


\end{document}